\documentclass{article}

\usepackage{arxiv}

\usepackage[utf8]{inputenc} 
\usepackage[T1]{fontenc}    
\usepackage{hyperref}       
\usepackage{url}            
\usepackage{booktabs}       
\usepackage{amsfonts}       
\usepackage{nicefrac}       
\usepackage{microtype}      
\usepackage{lipsum}		
\usepackage{graphicx}
\usepackage{natbib}
\usepackage{doi}
\usepackage{amsmath}
\usepackage{amssymb}
\usepackage{enumitem}
\usepackage{amssymb}
\usepackage{amstext}
\usepackage{amsmath}
\usepackage{amsthm}
\usepackage{bm}
\usepackage{multicol}
\usepackage{pslatex}
\usepackage{apalike}
\usepackage{tikz}
\usepackage{caption}
\usepackage{breqn}
\usepackage{float}

\newcommand\numberthis{\addtocounter{equation}{1}\tag{\theequation}}

\title{Improving usual Naive Bayes classifier performances with Neural Naive Bayes based models}

\author{
\href{https://orcid.org/0000-0003-3595-0826}
{\includegraphics[scale=0.06]{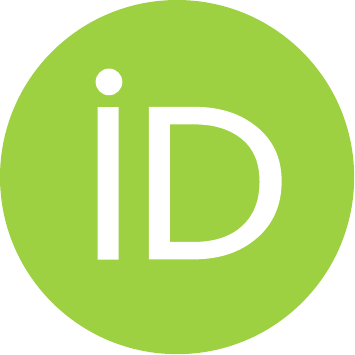}\hspace{1mm}
Elie~Azeraf}
\thanks{Elie Azeraf is also a member of Telecom SudParis, Institut Polytechnique de Paris.} \\
Watson Department \\
IBM France \\
Paris, France \\
\texttt{elie.azeraf@ibm.com} \\
\And
\href{https://orcid.org/0000-0002-7648-2515}
{\includegraphics[scale=0.06]{orcid.pdf}\hspace{1mm}
Emmanuel~Monfrini} \\
Telecom SudParis \\
Institut Polytechnique de Paris \\
Paris, France \\
\texttt{emmanul.monfrini@telecom-sudparis.eu} \\
\And
\href{https://orcid.org/0000-0002-1371-2627}
{\includegraphics[scale=0.06]{orcid.pdf}\hspace{1mm}
Wojciech~Pieczynski} \\
Telecom SudParis \\
Institut Polytechnique de Paris \\
Paris, France \\
\texttt{wojciech.pieczynski@telecom-sudparis.eu} \\
}

\date{}

\begin{document}
\maketitle

\begin{abstract}
Naive Bayes is a popular probabilistic model appreciated for its simplicity and interpretability. However, the usual form of the related classifier suffers from two major problems. First, as caring about the observations' law, it cannot consider complex features. Moreover, it considers the conditional independence of the observations given the hidden variable. This paper introduces the original Neural Naive Bayes, modeling the parameters of the classifier induced from the Naive Bayes with neural network functions. This allows to correct the first problem. We also introduce new Neural Pooled Markov Chain models, alleviating the independence condition. We empirically study the benefits of these models for Sentiment Analysis, dividing the error rate of the usual classifier by $4.5$ on the IMDB dataset with the FastText embedding.
\end{abstract}


\section{Introduction}
\label{sec:introduction}

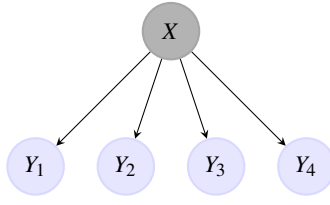
\begin{figure}[t]
\begin{center}
\begin{tikzpicture}
[font=\small, inner sep=0pt, visible/.style = {circle,draw = blue!15, fill = blue!10, thick,minimum size = 0.75cm, rounded corners}, hidden/.style = {circle,draw=black!35,fill=black!30,thick,minimum size=0.75cm, rounded corners}, scale = 0.6]
\node at (0,0) (x) [hidden] {$X$};

\node at (-3,-3) (y1) [visible]  {$Y_1$};
\draw [->, >=stealth] (x) to (y1);
                    
\node at (-1,-3) (y2) [visible]  {$Y_2$};
\draw [->, >=stealth] (x) to (y2);
                    
\node at (1,-3) (y3) [visible]  {$Y_3$};
\draw [->, >=stealth] (x) to (y3);

\node at (3,-3) (y4) [visible]  {$Y_4$};
\draw [->, >=stealth] (x) to (y4);
\end{tikzpicture}
\end{center}
\captionof{figure}{Probabilistic oriented graph of the Naive Bayes}
\label{fig_nb}
\end{figure}

We consider the hidden random variable $X$, taking its values in the discrete finite set $\Lambda_X = \{ \lambda_1, ..., \lambda_N \}$, and the observed random variables $Y_{1:T} = (Y_1, ..., Y_T), \forall t, Y_t \in \Omega_Y$. The Naive Bayes is a probabilistic model considering these variables. It is especially appreciated for its simplicity and interpretability, and is one of the most famous probabilistic graphical models (\cite{koller2009probabilistic,wainwright2008graphical}). This probabilistic model is defined with the following joint law:
\begin{align}
    p(X, Y_{1:T}) = p(X) \prod_{t = 1}^T p(Y_t | X).
\end{align}
It can be represented in figure \ref{fig_nb}.

We consider the Maximum A Posteriori (MAP) criterion. The Bayes classifier (\cite{devroye2013probabilistic,duda2006pattern,fukunaga2013introduction}) of the MAP can be written as follows, with the realization $y_{1:T}$ of $Y_{1:T}$ (we use the shortcut notation $p(X = x) = p(x)$:
\begin{align}
\phi(y_{1:T}) &= \arg \max_{\lambda_i \in \Lambda_X} p(X = \lambda_i | y_{1:T})
\end{align}

Therefore, the MAP classifier induced from the Naive Bayes is based on the posterior law, $\forall \lambda_i \in \lambda_X, p(X = \lambda_i | y_{1:T})$, usually written as follows:
\begin{align*}
    p(X = \lambda_i | y_{1:T}) &= \frac{p(X = \lambda_i, y_{1:T})}{\sum\limits_{\lambda_j \in \Lambda_X} p(X = \lambda_j, y_{1:T})} \\
    &= \frac{p(X = \lambda_i) \prod\limits_{t = 1}^T p(Y_t | X = \lambda_i)}{\sum\limits_{\lambda_j \in \Lambda_X} p(X = \lambda_j) \prod\limits_{t = 1}^T p(Y_t | X = \lambda_j)} \\
    &= \frac{\pi(i) \prod\limits_{t = 1}^T b_i^{(t)}(y_t) }{\sum\limits_{\lambda_j \in \Lambda_X} \pi(j) \prod\limits_{t = 1}^T b_j^{(t)}(y_t)}
    \label{nb_gen_clf}
    \numberthis
\end{align*}
with, $\forall \lambda_i \in \Lambda_X, y \in \Omega_Y, t \in \{ 1, ..., T \}$:
\begin{itemize}
    \item $\pi(i) = p(X = \lambda_i)$;
    \item $b_i^{(t)}(y) = p(Y_t = y | X = \lambda_i)$.
\end{itemize}

We consider for all this paper the stationary case for all models considered, i.e., the different parameters are not depending on time $t$. Thus, for the Naive Bayes, we set $b_i^{(t)}(y) = b_i(y)$.

The classifier induced from the Naive Bayes for classification with supervised learning always use the form (\ref{nb_gen_clf}), depending on the parameters $\pi$ and $b$ (\cite{jurafsky2000speech,ng2002discriminative,metsis2006spam,liu2013scalable}). It is applied in many applications, such as Sentiment Analysis or Text Classification (\cite{jurafsky2000speech,kim2006some,mccallum1998comparison}).

The classifier (\ref{nb_gen_clf}) is mainly criticized for two major drawbacks (\cite{ng2002discriminative,sutton2006introduction}). First, through the parameter $b$, it cares about the observations' law. This implies that one cannot consider complex features with the usual Naive Bayes classifier. Indeed, assuming that $Y_t \in \mathbb{R}^{d}$, and modeling $b$ by a Gaussian law, the number of parameters to learn is equal to:
\begin{align*}
    \underbrace{d}_{\text{for the mean}} + \underbrace{d^2}_{\text{for the covariance matrix}}
\end{align*}
If it is tractable for relatively small values of $d$, it quickly becomes intractable when $d$ increases. For example, with Natural Language Processing (NLP), it is common to convert words to numerical vectors of size $300$ (\cite{pennington2014glove}), $784$ (\cite{devlin2018bert}), or even $4096$ (\cite{akbik2018coling}), which becomes impossible to estimate. This process is called \textit{word embedding} and is mandatory to achieve relevant results. One can suppose the independence between the different components, also called features, of $Y_t$, resulting in estimating $d$ parameters for the covariance matrix, but this process achieves poor results. This "feature problem" also happens when the features belong to a discrete space, and approximation methods remain limited (\cite{brants2000tnt,azeraf2021highly}). 

Another main drawback of the Naive Bayes concerns the independence assumption of the observed random variables. It implies not considering the order of the different observations with the classifier induced from the stationary Naive Bayes model. 

In this paper, inspired by the writing of the classifier in (\cite{azeraf2021using}) and the Hidden Neural Markov Chain in (\cite{azeraf2021introducing}), we propose the Neural Naive Bayes. This model consists of defining the classifier induced from the Naive Bayes written in a discriminative manner with neural networks functions (\cite{lecun2015deep,Goodfellow-et-al-2016}). This neural model corrects the first default of the usual Naive Bayes classifier, as it can consider complex features of observations. Moreover, we propose the Neural Pooled Markov Chains, which are neural models assuming a conditional dependence of the observations given the hidden random variables, modeling them as a Markov chain. Therefore, they also correct the second drawback of the usual Naive Bayes classifier. Finally, we empirically study the contribution of these different innovations applied to the Sentiment Analysis task.

This paper is organized as follows. In the next section, we recall the classifier of the Naive Bayes written discriminatively, i.e., which does not use the observations' law, the Pooled Markov Chain (Pooled MC) and Pooled Markov Chain of order 2 (Pooled MC2) models and their classifiers. The third section presents our assumption to model the different parameters of these classifiers with neural network functions. Then, we empirically evaluate the contributions of our neural models applied to Sentiment Analysis. Conclusion and perspectives lie at the end of the paper.

To summarize our contributions, we present (1) three original neural models based on probabilistic models, and (2) we show their efficiency compared with the usual form of the Naive Bayes classifier for Sentiment Analysis.

\section{Naive Bayes and Pooled Markov Chains}

\subsection{The classifier induced from the Naive Bayes written discriminatively}

A classifier is considered "written generatively" if its form uses some $p(Y_A | X_B)$, with $A, B$ non-empty sets. If it is not written generatively, a classifier is written discriminatively. These notions rejoins the usual one about generative and discriminative classifiers (\cite{ng2002discriminative,jebara2012machine}). For example, (\ref{nb_gen_clf}) is the classifier induced from the Naive Bayes written generatively, as its form uses $b_i(y_t) = p(Y_t = y_t | X_t = \lambda_i)$.

(\cite{azeraf2021using}) presents how to write the classifier induced from the Naive Bayes written discriminatively:
\begin{align}
\phi^{NB}(y_{1:T}) = \arg \max_{\lambda_i \in \Lambda_X} \text{norm} \left( \pi(i)^{1 - T} \prod\limits_{t = 1}^T L_{y_t}(i) \right)
\label{nb_dis_clf}
\end{align}
with, $\forall \lambda_i \in \Lambda_X, t \in \{ 1, ..., T \}, y \in \Omega_Y$: 
\begin{align*}
    L_{y}(i) = p(X = \lambda_i | Y_t = y),
\end{align*}
and the norm function defined as, $\forall x \in \mathbb{R}^N$:
\begin{align*}
    \text{norm}: \mathbb{R} \rightarrow \mathbb{R}, x_i \rightarrow \frac{x_i}{\sum\limits_{j = 1}^N x_j}.
\end{align*}

\subsection{Pooled Markov Chain}

\begin{figure}[t]
\begin{center}
\begin{tikzpicture}
[font=\small, inner sep=0pt, visible/.style = {circle,draw = blue!15, fill = blue!10, thick,minimum size = 0.75cm, rounded corners}, hidden/.style = {circle,draw=black!35,fill=black!30,thick,minimum size=0.75cm, rounded corners}, scale = 0.6]
\node at (0,0) (x) [hidden] {$X$};

\node at (-3,-3) (y1) [visible]  {$Y_1$};
\draw [->, >=stealth] (x) to (y1);
                    
\node at (-1,-3) (y2) [visible]  {$Y_2$};
\draw [->, >=stealth] (x) to (y2);
                    
\node at (1,-3) (y3) [visible]  {$Y_3$};
\draw [->, >=stealth] (x) to (y3);

\node at (3,-3) (y4) [visible]  {$Y_4$};
\draw [->, >=stealth] (x) to (y4);

\draw [->, >=stealth] (y1) to (y2);
\draw [->, >=stealth] (y2) to (y3);
\draw [->, >=stealth] (y3) to (y4);
\end{tikzpicture}
\end{center}
\captionof{figure}{Probabilistic oriented graph of the Pooled Markov Chain}
\label{fig_pooled_mc}
\end{figure}
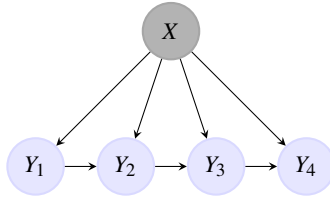

We introduce the Pooled Markov Chain model, considering the same random variables as the Naive Bayes. It is defined with the following joint law:
\begin{align}
    p(X, Y_{1:T}) = p(X) p(Y_1 | X) \prod_{t = 1}^{T - 1} p(Y_{t + 1} | X, Y_t)
    \label{pooledmc_joint_law}
\end{align}
This model extends the Naive Bayes, supposing that the observed random variables follow a Markov chain given the hidden variable, while the Naive Bayes supposes the independence.
It is represented in figure \ref{fig_pooled_mc}.

The classifier induced from the Pooled MC written discriminatively is defined as:
\begin{align}
\begin{split}
\phi^{MC}(y_{1:T}) = \arg \max_{\lambda_i \in \Lambda_X} \text{norm}
\left( L^{MC, 1}_{y_1}(i) \prod_{t = 1}^{T - 1} \frac{L^{MC, 2}_{y_t, y_{t + 1}}(i)}{L^{MC, 1}_{y_t}(i)} \right).
\label{pooledmc_dis_clf}
\end{split}
\end{align}
with the following parameter of the stationary Pooled MC, $\forall \lambda_i \in \Lambda_X, y_1, y_2 \in \Omega_Y$:
\begin{itemize}
    \item $L^{MC, 1}_{y_1}(i) = p(X = \lambda_i | Y_t = y_1)$;
    \item $L^{MC, 2}_{y_1, y_2}(i) = p(X = \lambda_i | Y_t = y_1, Y_{t + 1} = y_2)$.
\end{itemize}
The proof is given in the appendix.

\subsection{Pooled Markov Chain of order 2}

\begin{figure}[t]
\begin{center}
\begin{tikzpicture}
[font=\small, inner sep=0pt, visible/.style = {circle,draw = blue!15, fill = blue!10, thick,minimum size = 0.75cm, rounded corners}, hidden/.style = {circle,draw=black!35,fill=black!30,thick,minimum size=0.75cm, rounded corners}, scale = 0.6]
\node at (0,0) (x) [hidden] {$X$};

\node at (-3,-3) (y1) [visible]  {$Y_1$};
\draw [->, >=stealth] (x) to (y1);
                    
\node at (-1,-3) (y2) [visible]  {$Y_2$};
\draw [->, >=stealth] (x) to (y2);
                    
\node at (1,-3) (y3) [visible]  {$Y_3$};
\draw [->, >=stealth] (x) to (y3);

\node at (3,-3) (y4) [visible]  {$Y_4$};
\draw [->, >=stealth] (x) to (y4);

\draw [->, >=stealth] (y1) to (y2);
\draw [->, >=stealth] (y2) to (y3);
\draw [->, >=stealth] (y3) to (y4);

\draw [->,>=stealth] (y1.south) to [out=-30,in=-140] (y3.south);
\draw [->,>=stealth] (y2.south) to [out=-30,in=-140] (y4.south);
\end{tikzpicture}
\end{center}
\captionof{figure}{Probabilistic oriented graph of the Pooled Markov Chain of order 2}
\label{fig_pooled_mc2}
\end{figure}
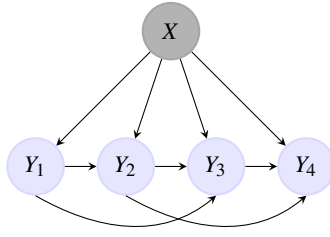

The Pooled Markov Chain of order 2, represented in figure \ref{fig_pooled_mc2} is defined with the joint law:
\begin{align}
\begin{split}
p(X, Y_{1:T}) = p(X) p(Y_1 | X) p(Y_2 | X, Y_1) \prod_{t = 1}^{T - 2} p(Y_{t + 2} | X, Y_t, Y_{t + 1})
\label{pooledmc2_joint_law}
\end{split}
\end{align}
This probabilistic model extends the two others, considering the observations follow a Markov chain of order 2 given the hidden variable.

The classifier induced from the Pooled MC2 written discriminatively is defined as:
\begin{align}
\begin{split}
\phi^{MC2}(y_{1:T}) = \arg \max_{\lambda_i \in \Lambda_X} \text{norm}
\left( L^{MC2, 1}_{y_1, y_2}(i) \prod_{t = 1}^{T - 1} \frac{L^{MC2, 2}_{y_t, y_{t + 1}, y_{t + 2}}(i)}{L^{MC2, 1}_{y_t, y_{t + 1}}(i)} \right).
\end{split}
\label{pooledmc2_dis_clf}
\end{align}
with the following parameter of the stationary Pooled MC2, $\forall \lambda_i \in \Lambda_X, y_1, y_2, y_3 \in \Omega_Y$:
\begin{itemize}
    \item $L^{MC2, 1}_{y_1, y_2}(i) = p(X = \lambda_i | Y_t = y_1, Y_{t + 1} = y_2)$;
    \item $L^{MC2, 2}_{y_1, y_2, y_3}(i) = p(X = \lambda_i | Y_t = y_1, Y_{t + 1} = y_2, Y_{t + 2} = y_3)$.
\end{itemize}
The proof is also given in the appendix.

\section{Neural Naive Bayes based models}

\subsection{Neural Naive Bayes}

We consider the classifier induced from the Naive Bayes written discriminatively (\ref{nb_dis_clf}). We assume the functions $\pi$ and $L$, allowing to define this classifiers, are strictly positive, and $\Omega_Y = \mathbb{R}^d, d \in \mathbb{N}^*$. (\ref{nb_dis_clf}) can be written as follows:
\begin{align*}
\phi^{NB}(y_{1:T}) &= \arg \max_{\lambda_i \in \Lambda_X} \text{norm} \left( \pi(i)^{1 - T} \prod\limits_{t = 1}^T L_{y_t}(i) \right) \\
&= \arg \max_{\lambda_i \in \Lambda_X} \text{norm} \left( \pi(i) \prod\limits_{t = 1}^T \frac{L_{y_t}(i)}{\pi(i)} \right) \\
&= \arg \max_{\lambda_i \in \Lambda_X} \text{softmax} \left( \log(\pi(i)) + \sum\limits_{t = 1}^T \log \left( \frac{L_{y_t}(i)}{\pi(i)} \right) \right).
\numberthis
\label{aaaa}
\end{align*}
with the softmax function defined as, $\forall x \in \mathbb{R}^N$:
\begin{align*}
    \text{softmax}: \mathbb{R} \rightarrow \mathbb{R}, x_i \rightarrow \frac{\exp(x_i)}{\sum\limits_{j = 1}^N \exp(x_j)}.
\end{align*}

Moreover, we suppose, $\forall i \in \{ 1, ..., N \}$, that the influence of $\log(\pi(i))$ is negligible in (\ref{aaaa}), which is true if the classes of $X$ have the same probability, or if $T$ is big enough.

We set $NN^{NB}$ a neural network function with $y_t$ as input and the output of size $N$. We define it as follows:
\begin{align*}
    NN^{NB}(y_t)_i = \log \left( \frac{L_{y_t}(i)}{\pi(i)} \right)
\end{align*}
with $NN^{NB}(y_t)_i$ the i-th component of the vector $NN^{NB}(y_t)$.

Therefore, (\ref{nb_dis_clf}) is approximated with:
\begin{align}
\phi^{NB}(y_{1:T}) \approx \arg \max_{\lambda_i \in \Lambda_X} \text{sotfmax} \left(  \sum\limits_{t = 1}^T NN^{NB}(y_t)_i \right)
\label{nnb_formula}
\end{align}
As (\ref{nnb_formula}) is the classifier induced from the Naive Bayes parametrized with neural networks, we call this application the Neural Naive Bayes.

\subsection{Neural Pooled Markov Chains}

We consider the Pooled MC model, and we assume the same hypothesis as above about the parameter functions. The classifier of the Pooled MC (\ref{pooledmc_dis_clf}) can be written:
\begin{align}
\phi^{MC}&(y_{1:T}) = \arg \max_{\lambda_i \in \Lambda_X} \text{softmax} \left( \log \left( L^{MC, 1}_{y_1}(i) \right) + \sum\limits_{t = 1}^{T - 1} \log \left( \frac{L^{MC, 2}_{y_t, y_{t + 1}}(i)}{L^{MC, 1}_{y_t}(i)} \right) \right).
\label{bbbb}
\end{align}

As above, we suppose that $\log \left( L^{MC, 1}_{y_1}(i) \right)$ is negligible. We define a neural network function, $NN^{MC}$, with the concatenation of $y_t$ and $y_{t + 1}$ as input, and an output of size $N$. It is used to model:
\begin{align*}
    NN^{MC}(y_t, y_{t + 1})_i = \log \left( \frac{L^{MC, 2}_{y_t, y_{t + 1}}(i)}{L^{MC, 1}_{y_t}(i)} \right)
\end{align*}

Therefore, (\ref{bbbb}) is approximated as follows:
\begin{align}
\phi^{MC}(y_{1:T}) \approx \arg &\max_{\lambda_i \in \Lambda_X} \text{sotfmax} \left(  \sum\limits_{t = 1}^{T - 1} NN^{MC}(y_t, y_{t + 1})_i \right)
\label{npooledmc_formula}
\end{align}

We called (\ref{npooledmc_formula}) the Neural Poooled Markov Chain (Neural Pooled MC) function.

In the same idea, the Neural Pooled Markov Chain of order 2 (Neural Pooled MC2) is defined with:
\begin{align}
\phi^{MC2}(y_{1:T}) \approx &\arg \max_{\lambda_i \in \Lambda_X} \text{sotfmax} \left(  \sum\limits_{t = 1}^{T - 2} NN^{MC2}(y_t, y_{t + 1}, y_{t + 2})_i \right)
\label{npooledmc2_formula}
\end{align}
with $NN^{MC2}$ a neural function having the concatenation of $y_t, y_{t + 1},$ and $y_{t + 2}$ as input, and an output of size $N$.

To go further, $\forall k \in \mathbb{N}^*,$ we introduce the Neural Pooled Markov Chain of order $k$ (Neural Pooled MC($k$)):
\begin{align}
\phi^{MCk}(y_{1:T}) &\approx \arg \max_{\lambda_i \in \Lambda_X} \text{sotfmax} \left(  \sum\limits_{t = 1}^{T - k} NN^{MCk}(y_t, y_{t + 1}, ..., y_{t + k})_i \right)
\label{npooledmck_formula}
\end{align}
with $NN^{MCk}$ a neural function having the concatenation of $y_t$ to $y_{t + k}$ as input, and an output of size $N$.

\section{Application to sentiment analysis}

\begin{table*}[t!]
    \centering
    \begin{tabular}{|l|c|c|c|c|}
        \hline
        {} & Usual Naive Bayes & Neural Naive Bayes & Neural Pooled MC & Neural Pooled MC2 \\
        \hline
        FastText & $49.49\%$ & $12.46\% \pm 0.19$ & $11.37\% \pm 0.24$ & $\bm{11.01\%} \pm 0.05$ \\
        \hline
        ExtVec & $46.64\%$ & $13.40\% \pm 0.34$ & $12.68\% \pm 0.17$ & $\bm{12.34\%} \pm 0.14$\\
        \hline
    \end{tabular}
    \captionof{table}{Accuracy error of the usual classifier induced from the Naive Bayes and our proposed neural models on IMDB dataset with FastText and ExtVec embeddings}
    \label{imdb_results}
\end{table*}

\begin{table*}[t!]
    \centering
    \begin{tabular}{|l|c|c|c|c|}
        \hline
        {} & Usual Naive Bayes & Neural Naive Bayes & Neural Pooled MC & Neural Pooled MC2 \\
        \hline
        FastText & $48.00\%$ & $15.32\% \pm 0.20$ & $14.50\% \pm 0.17$ & $\bm{14.10\%} \pm 0.26$ \\
        \hline
        ExtVec & $49.80\%$ & $17.22\% \pm 0.22$ & $16.14\% \pm 0.29$ & $\bm{15.88\%} \pm 0.14$ \\
        \hline
    \end{tabular}
    \captionof{table}{Accuracy error of the usual classifier induced from the Naive Bayes and our proposed neural models on SST-2 dataset with FastText and ExtVec embeddings}
    \label{sst2_results}
\end{table*}

\begin{table*}[t!]
    \centering
    \begin{tabular}{|l|c|c|c|c|}
        \hline
        {} & Usual Naive Bayes & Neural Naive Bayes & Neural Pooled MC & Neural Pooled MC2 \\
        \hline
        FastText & $60.38\%$ & $24.73\% \pm 0.16$ & $24.10\% \pm 0.17$ & $\bm{23.96\%} \pm 0.16$ \\
        \hline
        ExtVec & $60.38\%$ & $30.01\% \pm 0.22$ & $28.42\% \pm 0.18$ & $\bm{28.32\%} \pm 0.14$\\
        \hline
    \end{tabular}
    \captionof{table}{Accuracy error of the usual classifier induced from the Naive Bayes and our proposed neural models on TweetEval dataset with FastText and ExtVec embeddings}
    \label{tweet_results}
\end{table*}

Sentiment Analysis is a NLP task consisting of predicting the sentiment of a given text. In this section, we are going to apply the usual classifier induced from the Naive Bayes (\ref{nb_gen_clf}), the Neural Naive Bayes (\ref{nnb_formula}), the Neural Pooled MC (\ref{npooledmc_formula}), and the Neural Pooled MC2 (\ref{npooledmc2_formula}) for Sentiment Analysis. The goal is to observe the improvement brought by our new classifiers compared to the one usually used.

We use two different embedding methods: FastText (\cite{bojanowski2017enriching}) and ExtVec (\cite{komninos2016dependency}), allowing to convert each word in a given sentence to a vector of size $300$. Therefore, $\Omega_Y = \mathbb{R}^{300}$ is this use-case.

Every parameter of neural models, $NN^{NB}, NN^{MC}, NN^{MC2}$, is a feedforward neural network with the adapted input size, a hidden layer of size $64$ followed by the ReLU (\cite{nair2010rectified}) activation function, and an output layer of size $N$, the latter depending on the dataset.

\subsection{Dataset description}

We use three datasets for our experiments:
\begin{itemize}
    \item the IMDB dataset (\cite{maas-EtAl:2011:ACL-HLT2011}), composed of movie reviews, with a train set of $25000$ texts and a test set of the same size, with $\Lambda_X = \{ \textit{Positive}, \textit{Negative} \}$.
    \item the SST-2 dataset (\cite{socher2013recursive}) from GLUE \cite{wang2019glue}, also composed of movie reviews, with a train set of $67349$ texts, a test set of $1821$ texts, and a validation set of $872$ texts, with $\Lambda_X = \{ \textit{Positive}, \textit{Negative} \}$; 
    \item the TweetEval emotion dataset (\cite{barbieri2020tweeteval,mohammad2018semeval}), composed with Twitter data, with a train set of $3257$ texts, a test set of $1421$ texts, and a validation set of $374$ texts, with $\Lambda_X = \{ \textit{Anger}, \textit{Joy}, \textit{Optimism}, \textit{Sadness} \}$; 
\end{itemize}
All these datasets are freely available with the datasets library (\cite{lhoest2021datasets}).

\subsection{Parameter learning and implementation details}

On the one hand, the parameter $\pi$ and $b$ from (\ref{nb_gen_clf}) are learned with maximum likelihood estimation. For $\pi$, it consists of estimating it by counting the frequencies of the different patterns:
\begin{align*}
    \pi(i) = \frac{N_i}{\sum\limits_j N_j}
\end{align*}
with $N_i$ the number of times $X = \lambda_i$ in the training set. We model $b$ with a gaussian law. As the estimation of the covariance matrix of size $300 \times 300$ for each class is intractable, we select the vector's mean as the observation of each word. It results to have to estimate a mean and a variance of size $1$ for each class. Therefore, for each $\lambda_i \in \Lambda_X, b_i \sim \mathcal{N}(\mu_i, \sigma_i)$, with $\mu_i$ and $\sigma_i$ estimated by maximum likelihood.

About the parameters $NN^{NB}, NN^{MC},$ and $NN^{MC2}$, they are estimated with stochastic gradient descent algorithm using Adam (\cite{kingma2014adam}) optimizer with back-propagation (\cite{rumelhart1986learning,lecun1989backpropagation}). We minimize the Cross-Entropy loss function with a learning rate of $0.001$ and a mini-batch size of $64$. For the IMDB dataset, where the validation set is not directly given, we construct it with $25\%$ of the train set, drawn randomly. For each experiment, we train our model for $5$ epochs, and we keep the one achieving the best score on the validation set for testing. As it is a stochastic algorithm, every experiment with a Neural Naive Bayes based model is realized five times, and we report the mean and the $95\%$ confidence interval.

About the implementation details, all the codes are written in python. We use the Flair (\cite{akbik2019flair}) library for the word embedding methods, and PyTorch (\cite{paszke2019pytorch}) for neural functions and automatic differentiation. All experiments are realized with a CPU having 16Go RAM.

\subsection{Results}

We apply the usual classifier induced from Naive Bayes, the Neural Naive Bayes, the Neural Pooled MC, and the Neural Pooled MC2 to the three datasets. The accuracy errors are available in Table \ref{imdb_results}, \ref{sst2_results}, and \ref{tweet_results}, for the IMDB, SST-2, and TweetEval datasets, respectively.

As expected, one can observe an important error reduction with the neural models related to the usual classifier induced from the Naive Bayes. Indeed, for example, with FastText embedding on the IMDB dataset, this error is divided by about $4.5$ with the Neural Pooled MC2. Thus, it confirms the importance of considering complex features and shows the potential of our neural models for Sentiment Analysis.

Moreover, as one can observe in Table \ref{imdb_results}, \ref{sst2_results}, and \ref{tweet_results}, increasing the Markov chain's order allows to improve results. It shows the effect of alleviating the conditional independence condition. This observation is expected as the model becomes more complex and extended. Indeed, for each $k \in \mathbb{N}$, the Pooled Markov Chain of order $k$ is a particular case of the Pooled Markov Chain of order $k + 1$. To observe the limits of this case, we also apply the Neural Pooled MC(k), with $k \in \{ 3, 5, 7, 10 \}$. All these models achieve slightly equivalent results as the Neural Pooled MC2, without significant improvements, showing the empirical limits of this method.

\section{Conclusion and perspectives}

In this paper, we present the original Neural Naive Bayes and Neural Pooled Markov Chain models. These models significantly improve the usual Naive Bayes classifier's performances, considering complex observations' features without constraints, modeling their parameters with Neural Network functions, and alleviating the conditional independence hypothesis. They allow to combine performance and simplicity, and present a original way to combine neural networks with probabilistic models.

Moreover, these new neural models can be viewed as original pooling methods for textual data. Indeed, starting from word embedding methods, a usual way for text embedding consists of summing the embedding of the words of the text. Our methods, which benefit from being computationally light, propose a new way to do this document embedding process and can be included in any neural architecture. Therefore, going further in this direction can be a promising perspective.

\bibliographystyle{unsrtnat}
\bibliography{references}

\section*{APPENDIX}

\subsection*{Proof of the classifier induced from the Pooled MC written discriminatively}

The joint law of the Pooled MC is given with (\ref{pooledmc_joint_law}). For all $\lambda_i \in \Lambda_X$ and the realization $Y_{1:T} = y_{1:T}$, it can be written:
\begin{align*}
    p(X = \lambda_i, y_{1:T}) &= p(X = \lambda_i) p(y_1 | X = \lambda_i) \prod_{t = 1}^{T - 1} p(y_{t + 1} | X = \lambda_i, y_t) \\
    &= p(X = \lambda_i) \frac{p(X = \lambda_i, y_1)}{p(X = \lambda_i)} \prod_{t = 1}^{T - 1} \frac{p(X = \lambda_i, y_t, y_{t + 1})}{p(X = \lambda_i, y_t)} \\
    &= p(y_1) p(X = \lambda_i | y_1) \prod_{t = 1}^{T - 1} \frac{p(y_t) p(y_{t + 1} | y_t) p(X = \lambda_i | y_t, y_{t + 1})}{p(y_t) p(X = \lambda_i | y_t)} \\
    &= p(y_1) \prod_{t = 1}^{T - 1} p(y_{t + 1} | y_t) L^{MC, 1}_{y_1}(i) \prod_{t = 1}^{T - 1} \frac{L^{MC, 2}_{y_t, y_{t + 1}}(i)}{L^{MC, 1}_{y_t}(i)}
\end{align*}

Therefore, the posterior law of the Pooled MC can be written using the Bayes rule:
\begin{align}
    p(X = \lambda_i | y_{1:T}) = \frac{L^{MC, 1}_{y_1}(i) \prod\limits_{t = 1}^{T - 1} \frac{L^{MC, 2}_{y_t, y_{t + 1}}(i)}{L^{MC, 1}_{y_t}(i)}}{\sum\limits_{\lambda_j \in \Lambda_X} L^{MC, 1}_{y_1}(j) \prod\limits_{t = 1}^{T - 1} \frac{L^{MC, 2}_{y_t, y_{t + 1}}(j)}{L^{MC, 1}_{y_t}(j)} }
    \label{cccc}
\end{align}

Thus, (\ref{cccc}) allows to define the Bayes classifier with the MAP criterion of the Pooled MC as (\ref{pooledmc_dis_clf}), ending the proof.

\subsection*{Proof of the classifier induced from the Pooled MC2 written discriminatively}

The joint law of the Pooled MC2 is given with (\ref{pooledmc2_joint_law}). For all $\lambda_i \in \Lambda_X$ and the realization $Y_{1:T} = y_{1:T}$, it can be written:
\begin{align*}
    p(X = \lambda_i, y_{1:T}) &= p(X = \lambda_i) p(y_1 | X = \lambda_i) p(y_2 | X = \lambda_i, y_1) \prod_{t = 1}^{T - 2} p(y_{t + 2} | X = \lambda_i, y_t, y_{t + 1}) \\
    &= p(X = \lambda_i) \frac{p(X = \lambda_i, y_1)}{p(X = \lambda_i)} \frac{p(X = \lambda_i, y_1, y_2)}{p(X = \lambda_i, y_1)} \prod_{t = 1}^{T - 2} \frac{p(X = \lambda_i, y_t, y_{t + 1}, y_{t + 2})}{p(X = \lambda_i, y_t, y_{t + 1})} \\
    &= p(y_1, y_2) p(X = \lambda_i | y_1, y_2) \prod_{t = 1}^{T - 2} \frac{p(y_{t + 2} | y_t, y_{t + 1}) p(X = \lambda_i | y_t, y_{t + 1}, y_{t + 2})}{p(X = \lambda_i | y_t, y_{t + 1})} \\
    &= p(y_1, y_2) \prod_{t = 1}^{T - 2} p(y_{t + 2} | y_t, y_{t + 1}) L^{MC2, 1}_{y_1, y_2}(i) \prod_{t = 1}^{T - 2} \frac{L^{MC2, 2}_{y_t, y_{t + 1}, y_{t + 2}}(i)}{L^{MC, 1}_{y_t, y_{t + 1}}(i)}
    \numberthis
    \label{eeee}
\end{align*}

Therefore, the posterior law of the Pooled MC2 can be written using the Bayes rule and (\ref{eeee}):
\begin{align}
    p(X = \lambda_i | y_{1:T}) = \frac{L^{MC2, 1}_{y_1, y_2}(i) \prod_{t = 1}^{T - 2} \frac{L^{MC2, 2}_{y_t, y_{t + 1}, y_{t + 2}}(i)}{L^{MC, 1}_{y_t, y_{t + 1}}(i)}}{\sum\limits_{\lambda_j \in \Lambda_X} L^{MC2, 1}_{y_1, y_2}(j) \prod_{t = 1}^{T - 2} \frac{L^{MC2, 2}_{y_t, y_{t + 1}, y_{t + 2}}(j)}{L^{MC, 1}_{y_t, y_{t + 1}}(j)} }
    \label{dddd}
\end{align}

Thus, (\ref{dddd}) allows to define the Bayes classifier with the MAP criterion of the Pooled MC2 as (\ref{pooledmc2_dis_clf}), ending the proof.

\end{document}